\documentclass[10pt,twocolumn,letterpaper]{article}

\usepackage[pagenumbers]{wacv} 

\usepackage{graphicx}
\usepackage{amsmath}
\usepackage{amssymb}
\usepackage{booktabs}

\usepackage[dvipsnames]{xcolor}
\definecolor{cvprblue}{rgb}{0.21,0.49,0.74}
\definecolor{nicepink}{RGB}{255,70,70}
\usepackage[pagebackref,breaklinks,colorlinks,linkcolor=nicepink,citecolor=cvprblue]{hyperref}

\usepackage{array}

\usepackage{multirow}
\usepackage{subcaption}
\usepackage{orcidlink}

\usepackage[capitalize]{cleveref}
\crefname{section}{Sec.}{Secs.}
\Crefname{section}{Section}{Sections}
\Crefname{table}{Table}{Tables}
\crefname{table}{Tab.}{Tabs.}

\def\wacvPaperID{688} 
\def\confName{WACV}
\def\confYear{2025}

\begin{document}

\title{Optimizing Retinal Prosthetic Stimuli with \\ Conditional Invertible Neural Networks}

\author{Yuli Wu$^{1}\,$\orcidlink{0000-0002-6216-4911} \quad Julian Wittmann$^{1}\,$\orcidlink{0009-0003-3043-3843} \quad Peter Walter$^{2}\,$\orcidlink{0000-0001-8745-6593} \quad Johannes Stegmaier$^{1}\,$\orcidlink{0000-0003-4072-3759} \vspace{0.3em} \\
{$^1$ Institute of Imaging \& Computer Vision, RWTH Aachen University, Germany} \vspace{0.1em}\\
{$^2$ Department of Ophthalmology, RWTH Aachen University, Germany}\vspace{0.1em}  \\
{\tt\normalsize  \{yuli.wu,johannes.stegmaier\}@lfb.rwth-aachen.de}
}

\maketitle

\begin{abstract}
Implantable retinal prostheses offer a promising solution to restore partial vision by circumventing damaged photoreceptor cells in the retina and directly stimulating the remaining functional retinal cells. However, the information transmission between the camera and retinal cells is often limited by the low resolution of the electrode array and the lack of specificity for different ganglion cell types, resulting in suboptimal stimulations.
In this work, we propose to utilize normalizing flow-based conditional invertible neural networks to optimize retinal implant stimulation in an unsupervised manner.
The inherent invertibility of these networks allows us to use them as a surrogate for the computational model of the visual system, while also encoding input camera signals into optimized electrical stimuli on the electrode array.
Compared to other methods, such as trivial downsampling, linear models, and feed-forward convolutional neural networks, the flow-based invertible neural network and its conditional extension yield better visual reconstruction qualities \textit{w.r.t.} various metrics using a physiologically validated simulation tool. 
\end{abstract}

\section{Introduction}\label{sec:intro}

\begin{figure}[b]
\centering
\begin{subfigure}[t]{.25\textwidth}
 \hspace{-0.4cm}
  \includegraphics[width=\textwidth]{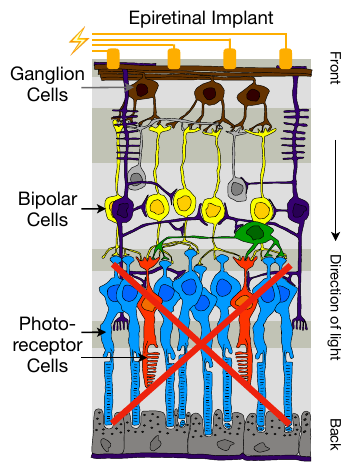}
  \caption{}
  \label{fig:ret_left}
\end{subfigure}
\begin{subfigure}[t]{0.15\textwidth}
\centering
  \includegraphics[width=0.9\textwidth]{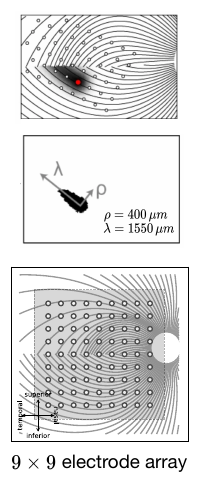}
  \caption{}
  \label{fig:ret_right}
\end{subfigure}
\caption{(a) Anatomy of the retina with an electrical epiretinal implant. Modified based on \textit{Retinal Implant} by Mbuerki under CC BY-SA 3.0 \cite{wiki:retinal}.  (b) Electrode activation with the axon map model (top, by Beyeler \etal \cite{beyeler2019model} under CC BY 4.0) and a retinal implant with a $9\times9$ electrode grid on the axon map (bottom, generated with \texttt{pulse2percept} \cite{michael_beyeler-proc-scipy-2017} ).}
\label{fig:ret}
\end{figure}

\begin{figure}[t]
    \centering
    \includegraphics[width=\linewidth]{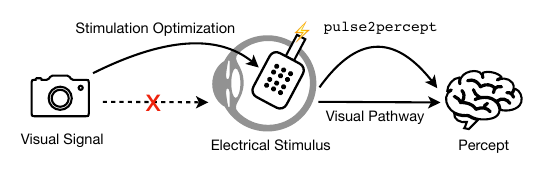}
    \caption{A schematic overview of the stimulation optimization for retinal prostheses. The visual signals are converted directly into electrical signals on the electrodes through an optimization process, constrained by a simulation model of the visual pathway.}
    \label{fig:overview}
\end{figure}

Retinal implants \cite{weiland2005retinal,zrenner2013fighting} aim to restore partial vision to individuals with certain types of visual impairments, particularly those caused by diseases like retinitis pigmentosa (RP) or age-related macular degeneration (AMD)  \cite{lim2012age,wong2014global,santos1997preservation,hartong2006retinitis,world2019world}. 
The visual signals captured by an external camera are transmitted to the electrical signals on the implanted electrode array, which bypass damaged photoreceptor cells in the retina and directly stimulate the remaining functional retinal cells (\Cref{fig:ret_left}). 
Various approaches exist to implant the prosthesis to different locations, such as epiretinal (facing the ganglion cells), subretinal (behind the photoreceptors), and suprachoroidal (on top of the choroid). 
Since we utilize the Axon Map Model (\Cref{fig:ret_right}) as the simulation model from the stimulus to the percept, which is based on the epiretinal implant Argus I/II devices~\cite{luo2016argus,Cruz19}, we limit our findings for epiretinal implants \cite{walter2005epiretinal}, such as Argus I/II~\cite{luo2016argus,Cruz19}, EPIRET3 \cite{mokwa2008intraocular,roessler2009implantation}, and VLARS \cite{waschkowski2014development,lohmann2019very}. 
However, the limited resolution and lack of specificity for different ganglion cell types (such as ON and OFF cells) necessitate an advanced algorithm for stimulation optimization \cite{sanes2015types,guo2013cell,jepson2013focal,lohler2023cell}.
The retinal prosthetic stimulation optimization can be considered as an inverse problem, where the optimized stimulus is achieved with the inverse transformation of the biologically fixed visual pathway (\Cref{fig:overview}). With the help of powerful machine learning and deep learning techniques \cite{beyeler2022towards,beech2024deep}, the optimization tasks can be solved, \eg, using an evolutionary heuristic algorithm \cite{romeni2021machine}, a reconstruction decoder \cite{van2022end} or reinforcement learning \cite{kuccukouglu2022optimization}.
Based on a physiologically validated simulation tool, \texttt{pulse2percept} \cite{michael_beyeler-proc-scipy-2017}, a line of deep learning-based approaches are proposed to learn an inverse model \cite{relic2022deep,wu2023bmt}, to learn a hybrid autoencoder \cite{granley2022hybrid}, or to learn an optimization encoder with a surrogate visual pathway simulation model in an end-to-end manner \cite{wu2023embc}. 
Besides, preference-based human-in-the-loop Bayesian optimization algorithms are proposed \cite{fauvel2022human,granley2023human}, which facilitate determining the patient-individual parameters of an \textit{in silico} retinal prosthetic computational model.

As one of the notable early papers of Invertible Neural Networks (INNs), Dinh \etal \cite{dinh2016density} propose to use INNs for density estimation tasks. 
Throughout a series of non-volume transformation processes, the \textit{normalizing} flow \cite{rezende2015variational} (total probability mass) is preserved, which ensures that the resulting distribution remains a valid probability distribution.
Conceptually, INNs are naturally suitable to construct an autoencoder \cite{kingma2013auto}, as the forward and backward direction of the INN should form a perfect encoder-decoder pair (but not always \cite{behrmann2021understanding}). Nguyen \etal \cite{nguyen2019training} train an INN autoencoder with an artificial bottleneck, created by setting several elements to zero before the inverse pass. Sorrenson and Draxler \etal \cite{sorrenson2024lifting} train autoencoders with likelihood on novel flow architectures that do not rely on coupling and are not dimension preserving. Furthermore, Ardizzone \etal first propose an INN variant to solve inverse problems with the maximum mean discrepancy loss (INN-MMD) \cite{ardizzone2018analyzing} and an extension to INNs with explicit controls, namely Conditional Invertible Neural Networks (cINNs) \cite{ardizzone2019guided,ardizzone2021conditional}. Analogous to other generative models, cINNs can also be applied to unsupervised data synthesis, such as \cite{dreher2023unsupervised}. 
Zhou \etal \cite{zhou2022neural} apply flow-based invertible generative models to neural encoding and decoding, transforming natural images and corresponding neural spike trains recorded from salamander retinal ganglion cells (RGC) \cite{onken2016using} bidirectionally. One key methodological distinction in our approach is that we directly train the \textit{stimulus-percept} pairs using a physiologically validated simulation tool \cite{michael_beyeler-proc-scipy-2017}. In contrast, the neural signals (spike trains) in \cite{zhou2022neural} are trained to reconstruct visual signals without a strong constraint in the decoder in the salamander RGC dataset \cite{onken2016using}.

\begin{figure}[t]
    \centering
\begin{subfigure}[t]{.47\textwidth}
    \includegraphics[width=\linewidth]{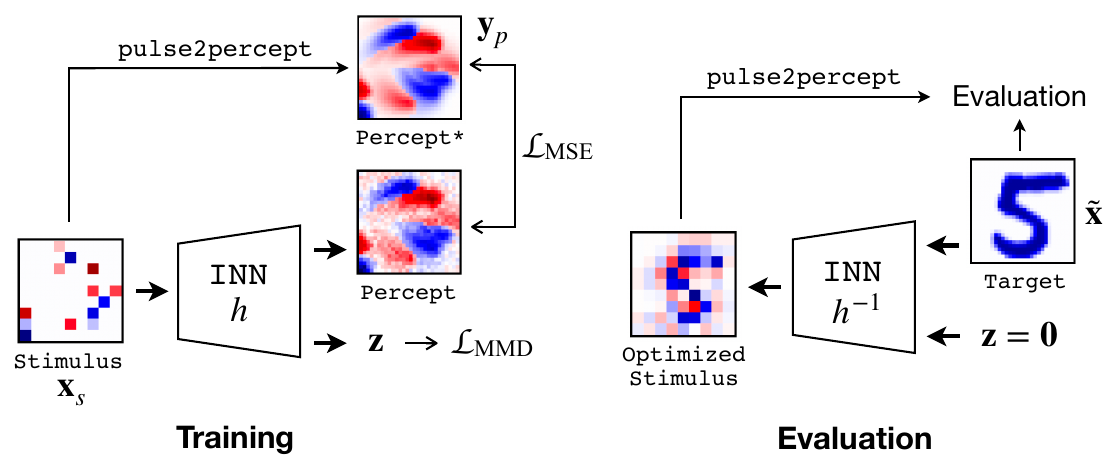}
    \caption{INN-MMD\\}
    \label{fig:2left}
\end{subfigure}\\\vspace{0.5cm}
\begin{subfigure}[t]{.47\textwidth}
    \includegraphics[width=\linewidth]{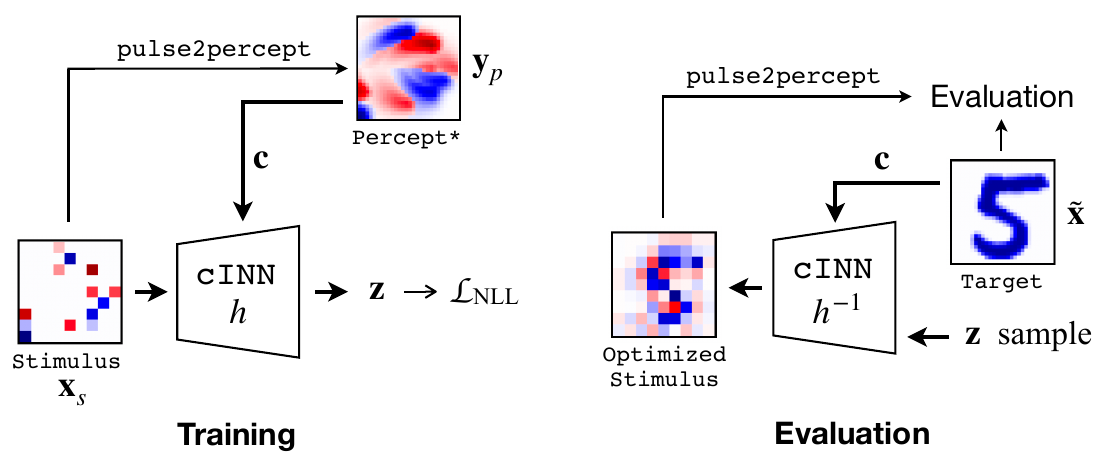}
    \caption{cINN}
    \label{fig:2right}
\end{subfigure}
\caption{Training and evaluation pipelines for the stimulation optimization with (a) invertible neural networks with maximum mean discrepancy loss (INN-MMD) \cite{ardizzone2018analyzing,ardizzone2019guided} and (b) conditional invertible neural networks (cINN) \cite{ardizzone2021conditional}.}
\label{fig:2}
\end{figure}

In this work, we follow an INN \cite{ardizzone2018analyzing} and a cINN approach \cite{ardizzone2019guided,ardizzone2021conditional} to learn a computational model for the visual pathway (\Cref{fig:overview}) and automatically obtain the inverse mapping from \textit{percept} to \textit{stimulus} thanks to the intrinsic feature of invertibility. Furthermore, INNs are generative and can be implemented to become bijective to solve inverse problems \cite{ardizzone2018analyzing}. In addition, by imposing conditioning to the INN, the generated inverse input (\textit{optimized stimulus}, see \Cref{fig:2}) shall follow the desired constraint. We compare the reconstruction quality of the predicted percepts using cINNs to other approaches, including trivial downsampling, a linear model, feed-forward and invertible neural networks, and address the promising optimization for retinal prosthetic stimuli.

\section{Basics of Invertible Neural Network}\label{sec:inn}

Given a random variable $\mathbf{z}=h \left(\mathbf{x}\right)$ with a known and tractable density, \eg, a spherical multivariate normal distribution $\mathbf{z}\sim\pi \left(\mathbf{z}\right) = \mathcal{N}\left(\mathbf{z};\mathbf{0},\mathbf{I}\right)$, a data point $\mathbf{x}$ from an \textit{i.i.d.} dataset $\mathcal{D}$ can be generated by $\mathbf{x}=h^{-1}(\mathbf{z})$ for an invertible function $h$. Following the \textit{change of variables} theorem, the probability density function of such a model given a data point $\mathbf{x}$ can be written as:
\begin{equation}\label{eq:pdf1}
    p(\mathbf{x}) = \pi (h(\mathbf{x}))\left| \det \frac{\partial h(\mathbf{x})}{\partial \mathbf{x}^T}\right| = \pi (\mathbf{z}) \left| \det \mathbf{J}_h(\mathbf{x})\right| ,
\end{equation}
where $ \mathbf{J}_h(\mathbf{x}) = \partial h(\mathbf{x}) / \partial\mathbf{x}^T$ is the Jacobian matrix of $h$ at $\mathbf{x}$ and $\left| \det \mathbf{J}_h(\mathbf{x})\right|$ is the absolute value of the Jacobian determinant. Different names describe this approach from different perspectives of the likelihood (\Cref{eq:pdf1}): \textit{Invertible neural network} (INN) points out the intrinsic architecture design of $h$ shall be invertible; \textit{Normalizing flow} (NF) emphasizes that a data distribution of $\mathbf{x}$ is normalized into a simpler form of $\mathbf{z}$ by a series of transformations (flow); \textit{Non-volume preserving} (NVP, \cite{dinh2016density}) specifies that the scaling factor $\left| \det \mathbf{J}_h(\mathbf{x})\right|$ of volume elements under the flow transformation is not constrained to be $1$.

 With $\mathbf{x}$ and $\mathbf{z}$ sharing the identical dimensionality $\mathbf{x}, \mathbf{z}\in \mathbb{R}^n$ (\textit{cf.} \Cref{sec:mmdinn}), the normalizing flow $h$ can be trained by maximizing likelihood of $\mathbf{x}$, which is equivalent to minimizing the Negative Log-Likelihood (NLL) loss:
\begin{align}
    \mathcal{L}_{\mathrm{NLL}} &= -\log (p(\mathbf{x}))\\\label{eq:nll_z}
    &=  -\log \left( \pi (\mathbf{z}) \right)-\log \left|\det \mathbf{J}_h\left(\mathbf{x}\right)  \right| \\
    &=  -\log \left(\mathcal{N}\left( \mathbf{z};\mathbf{0},\mathbf{I}\right)\right)-\log \left|\det \mathbf{J}_h\left(\mathbf{x}\right)  \right| \\
    & = -\log \prod_{i=1}^{n} \left( \frac{1}{\sqrt{2\pi}} e^{ -\frac{z_i^2 }{2} } \right) -\log \left|\det \mathbf{J}_h\left(\mathbf{x}\right)  \right|  \\
    & = \frac{n}{2} \log(2\pi) + \frac{1}{2}\| \mathbf{z} \|_2^2 -\log \left|\det \mathbf{J}_h\left(\mathbf{x}\right)  \right|  \\
    &\simeq \frac{1}{2} \| h\left(\mathbf{x}\right)\|_2^2-\log \left|\det \mathbf{J}_h\left(\mathbf{x}\right)\right|.\label{eq:nll}
\end{align}

The first term in \Cref{eq:nll} can be easily calculated with the output of the forward normalizing flow. The second term is computationally expensive mainly due to the matrix determinant of the Jacobian. A line of INN architecture designs, \eg, additive coupling~\cite{dinh2014nice}, affine coupling~\cite{dinh2016density} and Glow coupling~\cite{kingma2018glow}, has been proposed to utilize an LU decomposition style architecture (\Cref{eq:lu}) to achieve the following properties. (1) Invertibility: The design of the coupling layer ensures that the transformation is invertible, which is crucial for computing the exact likelihood of the data under the transformed distribution. (2) Computational efficiency:
Both forward and inverse transformations are computationally efficient, which allows the model to scale to high-dimensional data. (3) Easy Jacobian: The Jacobian matrix of the transformation is triangular, making the computation of its determinant straightforward.
The coupling layer is structurally similar to the lifting scheme architecture~\cite{sweldens1998lifting} used in discrete wavelet transform. In an affine coupling layer~\cite{dinh2016density}, the input is split into two parts with $\mathbf{x} := [ \mathbf{x}_{A}, \mathbf{x}_{B}]$, only one part of which is applied with an affine transformation and the other remains unchanged:
\begin{align}
\begin{cases}
    \mathbf{z}'_{A} = \mathbf{x}_{A} \\
    \mathbf{z}'_{B} = \mathbf{x}_{B} \odot \exp(s(\mathbf{x}_{A})) + t(\mathbf{x}_{A})\,,
\end{cases}
\end{align}
where  \( \odot \) denotes the element-wise multiplication, while \( s(\cdot) \) and \( t(\cdot) \) are scale and translation functions, respectively. \(\mathbf{z}' := [ \mathbf{z}'_{A}, \mathbf{z}'_{B}]\) represents the intermediate output of the forward one-sided transformation. 
The invertibility of the first half of the affine coupling layer is straightforward and efficient due to its structure with:
\begin{align}
\begin{cases}
    \mathbf{x}_{A} = \mathbf{z}'_{A} \\
    \mathbf{x}_{B} = \left( \mathbf{z}'_{B} - t(\mathbf{x}_{A})  \right) \odot  \exp(-s(\mathbf{x}_{A}))  \,.
\end{cases}
\end{align}
Note that \(s(\cdot) \) and \( t(\cdot) \) are not required to be invertible themselves and can be any differentiable function (to facilitate gradient-based optimization) including neural networks.
The Jacobian matrix of this forward one-sided coupling transformation from $\mathbf{x}$ to $\mathbf{z}'$ is triangular, making its log-determinant easy to compute: 
\begin{align} 
\log \left|\det (\mathbf{J})\right| &= \log \left |{ \begin{array}{cccc} \mathbf{I} & \mathbf {0} \\ 
 \frac{\partial \mathbf{z}'_{B}}{ \partial {\mathbf{x}_A }} & \text {diag} \left ({\exp \left({s \left( \mathbf {x}_A \right) }\right)}\right) \\ \end{array}} \right|\\  \label{eq:lu}
&= \sum s({\mathbf {x}_A}) .
\end{align}
Additionally, in the subsequent layer, the second segment remains unchanged, whereas the first segment will be transformed with the intermediate output from the other \textit{side}: 
\begin{align}
\begin{cases}
    \mathbf{z}_{A} = \mathbf{z}'_{A} \odot \exp(s(\mathbf{z}'_{B})) + t(\mathbf{z}'_{B})\\
    \mathbf{z}_{B} = \mathbf{z}'_{B}\,.
\end{cases}
\end{align}
More details about the INN architectures used in this paper can be found in \Cref{sec:imp} and in \Cref{fig:a7}.

The original generative paradigm (\eg, RealNVP \cite{dinh2016density}) is not suitable for the application of retinal stimulus optimization, as the optimized stimulus shall be guided by the target visual signal as the reference. Therefore, we utilize two INN variants, namely INN-MMD \cite{ardizzone2018analyzing} with the combination of an unsupervised MMD loss (\Cref{sec:mmdinn}) and a supervised MSE loss, and cINN \cite{ardizzone2019guided,ardizzone2021conditional} with an explicit conditioning mechanism (\Cref{sec:cinn}).

\section{Methods}
\subsection{Stimulation Optimization}\label{sec:problem}
We begin by formulating the \textit{in silico} optimization problem for retinal prosthetic stimuli. Let ${f}$ represent the function of the visual pathway (\Cref{fig:overview}), which uses the electrical stimulus $\mathbf{x}_s$ from the implanted electrode array as input and produces the percept $\mathbf{y}$ as output. We assume that $f_{\phi}$ can simulate ${f}$, where $\phi$ denotes the parameters of the computational model, such as the Axon Map Model \cite{beyeler2019model}. The desired encoding function $g_{\theta}: \tilde{\mathbf{x}} \mapsto \mathbf{x}_s$ from the visual signal (\textit{target}) $\tilde{\mathbf{x}}$ to the stimulus ${\mathbf{x}_s}$ can be found with the following optimization problem: 
\begin{equation}
    \underset{\theta}{\mathrm{argmin}}\,\mathcal{L}_{\mathrm{eval}} \left( \tilde{\mathbf{x}}, f_{\phi}\left( g_{\theta} \left(\tilde{\mathbf{x}}\right)\right) \right) ,
\end{equation}
where $\mathcal{L}_{\mathrm{eval}}$ serves as the objective function. We employ a neural network to construct $g_{\theta}$ with learnable parameters $\theta$. In practice, we often train a differentiable neural network as a surrogate for $f_{\phi}$ (\eg, \cite{wu2023embc, granley2022hybrid}), enabling convenient end-to-end optimization for the encoder $g_{\theta}$.

\begin{figure}[t]
    \centering
    \includegraphics[width=\linewidth]{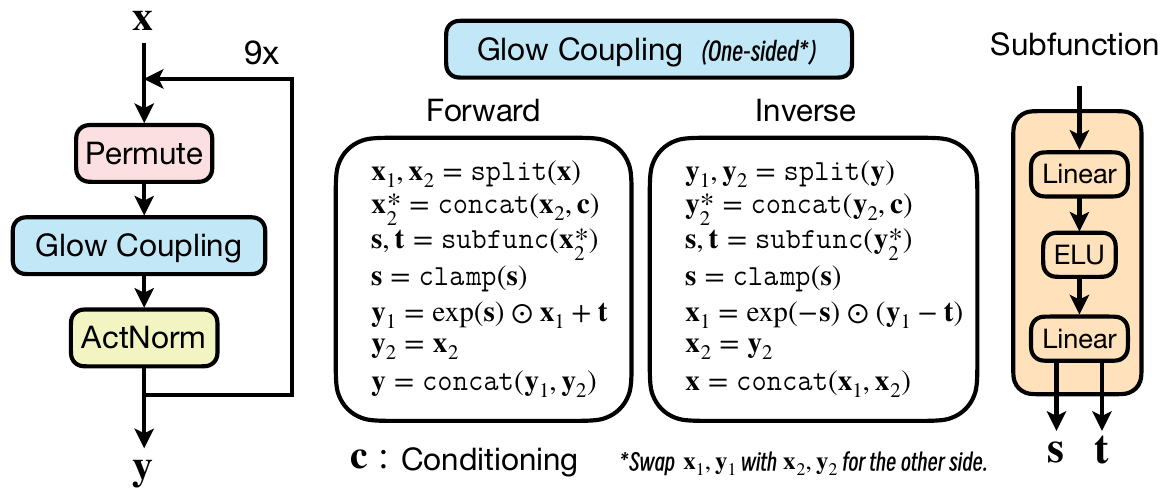}
    \caption{We follow \cite{ardizzone2019guided,ardizzone2021conditional} to design the INN and cINN architectures with Glow coupling layers \cite{kingma2018glow}. The conditional input $\mathbf{c}$ in both forward and inverse Glow coupling blocks is the only difference between the INN and the cINN approach. }
    \label{fig:a7}
\end{figure}

\subsection{Invertible Neural Network for Inverse Problems with Maximum Mean Discrepancy}\label{sec:mmdinn}

If a process $q: \mathbf{x}_D\in \mathbb{R}^D \mapsto \mathbf{y}_d\in \mathbb{R}^d$ is not bijective with $D>d$, we can introduce a latent random variable $\mathbf{z}_K\sim \pi \left(\mathbf{z}_K\right) =\mathcal{N}\left(\mathbf{z}_K;\mathbf{0},\mathbf{I}_K\right)$ with $K=D-d$, as proposed by Ardizzone \etal \cite{ardizzone2018analyzing}. We then construct an invertible process $h:\mathbf{x}_D\mapsto [\mathbf{y}_d,\mathbf{z}_K]$ with $h(\mathbf{x}_D)= [h_{\mathbf{y}_d}(\mathbf{x}_D), h_{\mathbf{z}_K}(\mathbf{x}_D)]$ and $h_{\mathbf{y}_d}(\mathbf{x}_D)\approx q(\mathbf{x}_D)$. Similar to \cite{ardizzone2018analyzing}, we apply a supervised Mean Squared Error (MSE) loss on $\mathbf{y}_d$ with $ \mathcal{L}_{\mathrm{MSE}} = \mathbb{E}[( \mathbf{y}_d - h_{\mathbf{y}_d}(\mathbf{x}_D))^2]$ and an unsupervised Maximum Mean Discrepancy (MMD) loss \cite{gretton2012kernel,gretton2006kernel} in terms of $\mathbf{x}_D$ and $\mathbf{z}_K$, respectively. 
The MMD loss enforces two distributions $p_{M}$ and $p_{M'}$ to be identical and is formulated as:
\begin{equation}
\begin{split}
    \mathcal{L}_{\mathrm{MMD}} (p_{M}, p_{M'})= & \\  \left(\mathbb{E}_{i,j}[k (\mu_i, \mu_j) \right.  
    & \left. - 2k (\mu_i, \mu_j')  + k (\mu_i', \mu_j')] \right) ^{\frac{1}{2}},
\end{split}
\end{equation}
where $\mathbb{E}$ denotes the expected value and $\{\mu\}, \{\mu'\}$ are two sets of samples according to the distributions $\mu \sim p_{M}$ and $\mu'\sim p_{M'}$, respectively.

\begin{figure}[t]
\centering
\begin{subfigure}[t]{.36\textwidth}
  \centering
  \includegraphics[width=\textwidth]{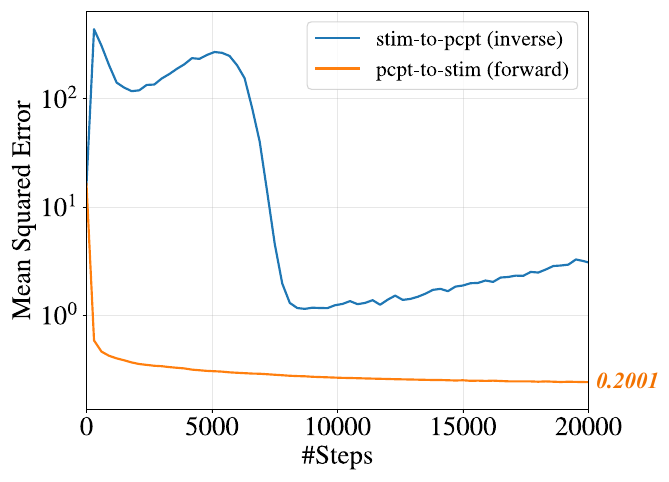}
  \caption{Forward process: \textit{percept-to-stimulus}.}
  \label{fig:loss_a}
\end{subfigure}\\
\par\bigskip
\begin{subfigure}[t]{.36\textwidth}
  \centering
  \includegraphics[width=\textwidth]{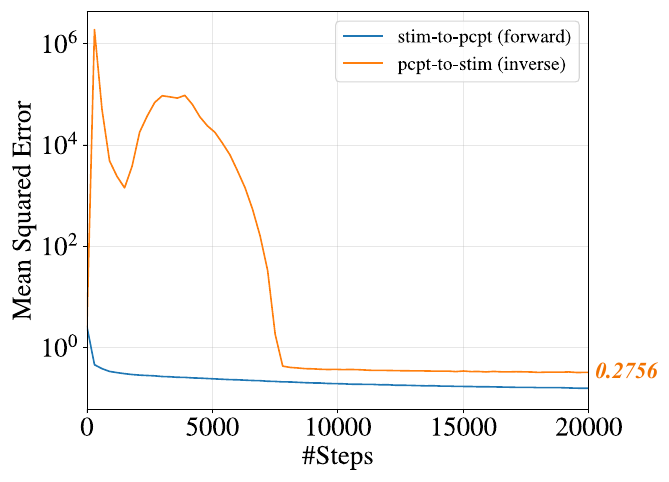}
  \caption{Forward process: \textit{stimulus-to-percept}.}
  \label{fig:loss_b}
\end{subfigure}
    \caption{MSE loss values of the INN-MMD training \cite{ardizzone2018analyzing} (a) for \textit{percept-to-stimulus} being the forward process and (b) for \textit{stimulus-to-percept} being the forward process. We observe that the MSE loss values of inverse processes are always larger than the supervised forward processes, which implies that the invertibility using a supervised INN is not guaranteed. We report the results of the setup (a) in \Cref{tab:1}, as the optimized stimulus is desired.}
\label{fig:a6}
\end{figure}

As suggested in \cite{ardizzone2018analyzing,tolstikhin2018wasserstein}, we utilize the radial basis function-based (RBF) rational quadratic kernel~\cite{muandet2017kernel} (also known as inverse multiquadric kernel):
\begin{equation}
    k (\mu, \mu') = \left( 1+\left\Vert (\mu-\mu') / \sigma \right\Vert_2^2\right) ^{-1}\,.
\end{equation} The MMD loss on $\mathbf{x}_D$ forces it to match the prior with $\mathcal{L}_\mathrm{MMD} (p(\mathbf{x}_D), p(h^{-1}([\mathbf{y}_d,\mathbf{z}_K])))$ and is optional. The MMD loss on $\mathbf{z}_K$ forces it to match the normal distribution with $\mathcal{L}_\mathrm{MMD} (p(h(\mathbf{x}_D)), p(\mathbf{y}_d)p(\mathbf{z}_K))$. 
Moreover, following the approach in \cite{ardizzone2018analyzing}, we pad $\mathbf{x}_D$ with zeros to match the required dimensionality when necessary. This allows us to train both directions as the forward pass, as illustrated in \Cref{fig:a6}.
Under the constraint of the invertible network architecture, both $h$ and $h^{-1}$ can be trained in both directions with a guaranteed invertible relation. 
Numerous invertible neural network architecture designs have been proposed to ensure that the inverse and the Jacobian determinant are easily computed, such as additive coupling layer \cite{dinh2014nice}, affine coupling layer \cite{dinh2016density} and Glow coupling layer \cite{kingma2018glow} (see \Cref{sec:inn}). The implementation details of the architecture in this work are visualized in \Cref{fig:a7}.

\begin{table}[b]
    \small
    \centering
    \caption{Summary of notations for retinal implants (\Cref{sec:problem}) in the context of invertible neural networks.}
    \begin{tabular}{lll}
    \toprule
    Notations    & INN-MMD \cite{ardizzone2018analyzing} & cINN \cite{ardizzone2019guided,ardizzone2021conditional} \\
    \midrule
     $f_\phi$      & \multicolumn{2}{c}{Forward network ($h$)}  \\
  $g_\theta$      & \multicolumn{2}{c}{Reverse network ($h^{-1}$)} \\
    $\mathbf{x}_s$       & \multicolumn{2}{c}{Forward input ($\mathbf{x}$ or $\mathbf{x}_D$)}   \\
    $\mathbf{\tilde{x}}$       & Reverse input &  Reverse conditioning   \\
    $\mathbf{y}$, $\mathbf{y}_d$    & Forward output (MSE) & Forward conditioning \\
    $\mathbf{z}$, $\mathbf{z}_K$      & Forward output (MMD) &  Forward output (NLL) \\
         \bottomrule
    \end{tabular}
    \label{tab:label_1}
\end{table}

\subsection{Conditional Invertible Neural Network}\label{sec:cinn}

Within the local coupling blocks~\cite{dinh2016density}, conditioning features $\mathbf{c}$ can be integrated into both directions of the network by concatenating them in the subfunctions with $\mathbf{z}=h(\mathbf{x};\mathbf{c})$ and $\mathbf{x}=h^{-1}(\mathbf{z};\mathbf{c})$, forming Conditional Invertible Neural Networks (cINNs)~\cite{ardizzone2019guided,ardizzone2021conditional}. This is similar to other conditional generative models, \eg, conditional variational auto-encoder (cVAE) \cite{sohn2015learning} and conditional generative adversarial network (cGAN) \cite{isola2017image}. 
Accordingly, the NLL loss for cINN maximum likelihood training is formulated as 
\begin{equation*}
    \mathcal{L}_{\mathrm{NLL}} \simeq \frac{1}{2} \| h(\mathbf{x}; \mathbf{c})\|_2^2-\log \left|\det \mathbf{J}_h(\mathbf{x})\right|\,.
\end{equation*}

As shown in \Cref{fig:2right}, we train the cINN by minimizing the NLL loss in an unsupervised manner and use the predicted percept during the training and the MNIST target during the evaluation as the conditioning. The same invertible neural network architecture is used as in the INN experiments. 
Unlike INN-MMD, which is constrained by a hybrid loss (combining supervised MSE and unsupervised MMD), cINN explicitly uses the conditioning as a reference to guide the \textit{stimulus} with the \textit{target}. 
A summary of notations for retinal implants from \Cref{sec:problem} is listed in \Cref{tab:label_1}, along with their corresponding contexts in INN-MMD and cINN.

\section{Experiments and Results}

\subsection{Axon Map Model}\label{sec:axon}
In our \textit{in silico} experiments, we create a virtual square retinal implant with a $9\times9$ electrode grid (\Cref{fig:ret_right}) and utilize the physiologically validated Axon Map Model \cite{beyeler2019model}, implemented in \texttt{pulse2percept} \cite{michael_beyeler-proc-scipy-2017}. The phosphenes elicited by the axons, somas, or dendrites of the retinal ganglion cells have different shapes, as demonstrated in 1999 by Greenberg \etal \cite{greenberg1999computational}. 
The recent Axon Map Model \cite{beyeler2019model} aligns with these findings by modelling the drawings from the subjects with retinal implants, which is one of the available models in \texttt{pulse2percept} \cite{michael_beyeler-proc-scipy-2017} along with such as Scoreboard Model \cite{beyeler2019model} and Biphasic Axon Map Model \cite{granley2021computational}.
The predicted percept intensity at location $(x,y)$ is calculated by:
\begin{equation}
    \begin{split}
    I(x,y) = \exp & \left( - \frac{(x-x_{\text{stim}})^2+(y-y_{\text{stim}})^2}{2\rho^2} \right. \\
    &\left. - \frac{(x-x_{\text{soma}})^2+(y-y_{\text{soma}})^2}{2\lambda^2} \right),
    \end{split}
\end{equation}
with exponential decays in terms of the distances to the stimulus at $(x_{\text{stim}},y_{\text{stim}})$ and to the elongated soma at $(x_{\text{soma}},y_{\text{soma}})$, controlled by the constants $\rho$ and $\lambda$, respectively (see \Cref{fig:ret_right}).
 In this paper, we set the constants to the realistic and challenging values $\rho = 400 \,\mu m$ and $\lambda = 1550\, \mu m$ for all experiments (\textit{cf.} \cite{beyeler2019model}).

\subsection{Datasets}
\begin{figure}[t]
    \centering
\begin{subfigure}[t]{.48\textwidth}
    \includegraphics[width=\linewidth]{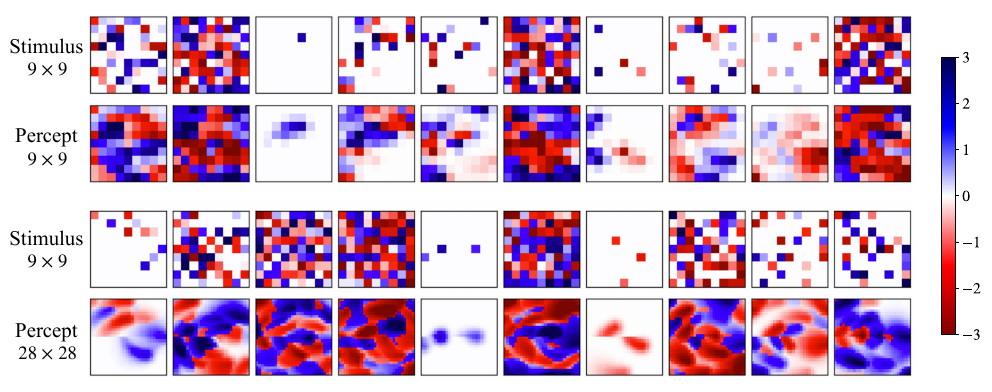}
    \caption{Examples of random samples from $9\times9$ stimuli to both $9\times9$ and $28\times28$ percepts using Axon Map Model \cite{beyeler2019model} implemented in \texttt{pulse2percept} \cite{michael_beyeler-proc-scipy-2017}.}
    \label{fig:a5}
\end{subfigure}\\
\begin{subfigure}[t]{.48\textwidth}
    \includegraphics[width=\linewidth]{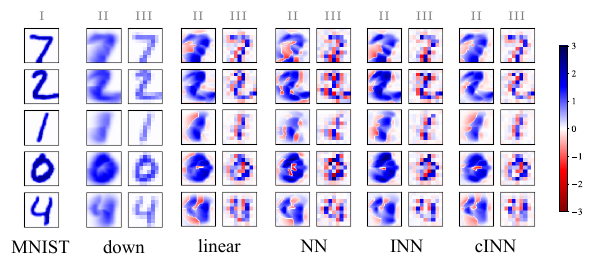}
    \caption{Visualization of the reconstruction quality with five approaches. (I) denotes the original MNIST targets. (II) denotes the predicted percepts of $28\times28$ pixels based on (III) optimized stimuli of $9\times9$ electrodes.}
    \label{fig:vis_b}
\end{subfigure}\\
\caption{Visualization during the training and the inference.}
\label{fig:vis}
\end{figure}

In order to achieve generalizability, models are trained using 10 million random \textit{stimulus-percept} pairs and then evaluated with the MNIST test set \cite{lecun2010mnist}. We conduct all experiments using a batch size of 1024.
The stimulus always has a resolution $9\times9$ pixels (\ie, the number of the electrodes) and we experiment with both $9\times9$ and $28\times28$ pixels for percepts and the MNIST targets. Each \textit{stimulus-percept} pair of the training set is generated by uniformly sampling in $[-3, +3]$ for each electrode. We select this intensity range to amplify the MSE values for particularly inaccurate predictions, when the difference between the predicted and ground-truth values exceeds 1.
Negative voltage values in this range are beneficial in inhibiting phosphene elongation, which effectively designates the corresponding electrodes as \textit{return} electrodes, as described in \cite{tong2020stimulation, wang2022electronic}.
We visualize stimuli and percepts with negative values in \Cref{fig:vis}, which do not result in any stimulated phosphenes though (similar to the neuron's refractory period).
Additionally, for each stimulus, only a random percentage of electrodes are activated while the rest are set to zero. Examples of \textit{stimulus-percept} pairs are showcased in \Cref{fig:a5}.

\begin{figure}[t]
\centering
\begin{subfigure}[t]{.34\textwidth}
  \centering
  \includegraphics[width=\textwidth]{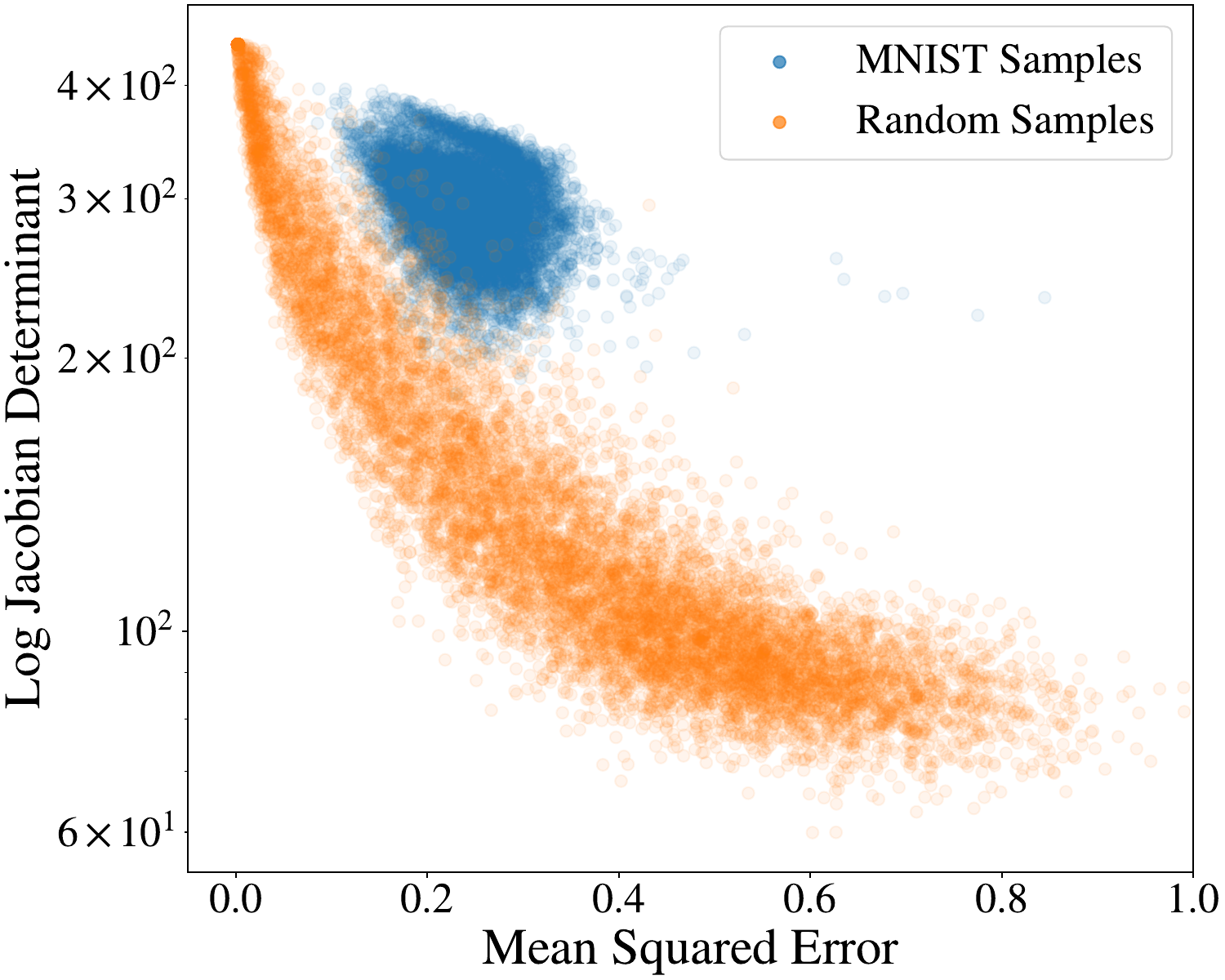}
  \caption{Correlation between log Jacobian determinant and MSE for MNIST and random samples.}
  \label{fig:likelihood1}
\end{subfigure}\\
\par\bigskip
\begin{subfigure}[t]{.34\textwidth}
  \centering
  \includegraphics[width=\textwidth]{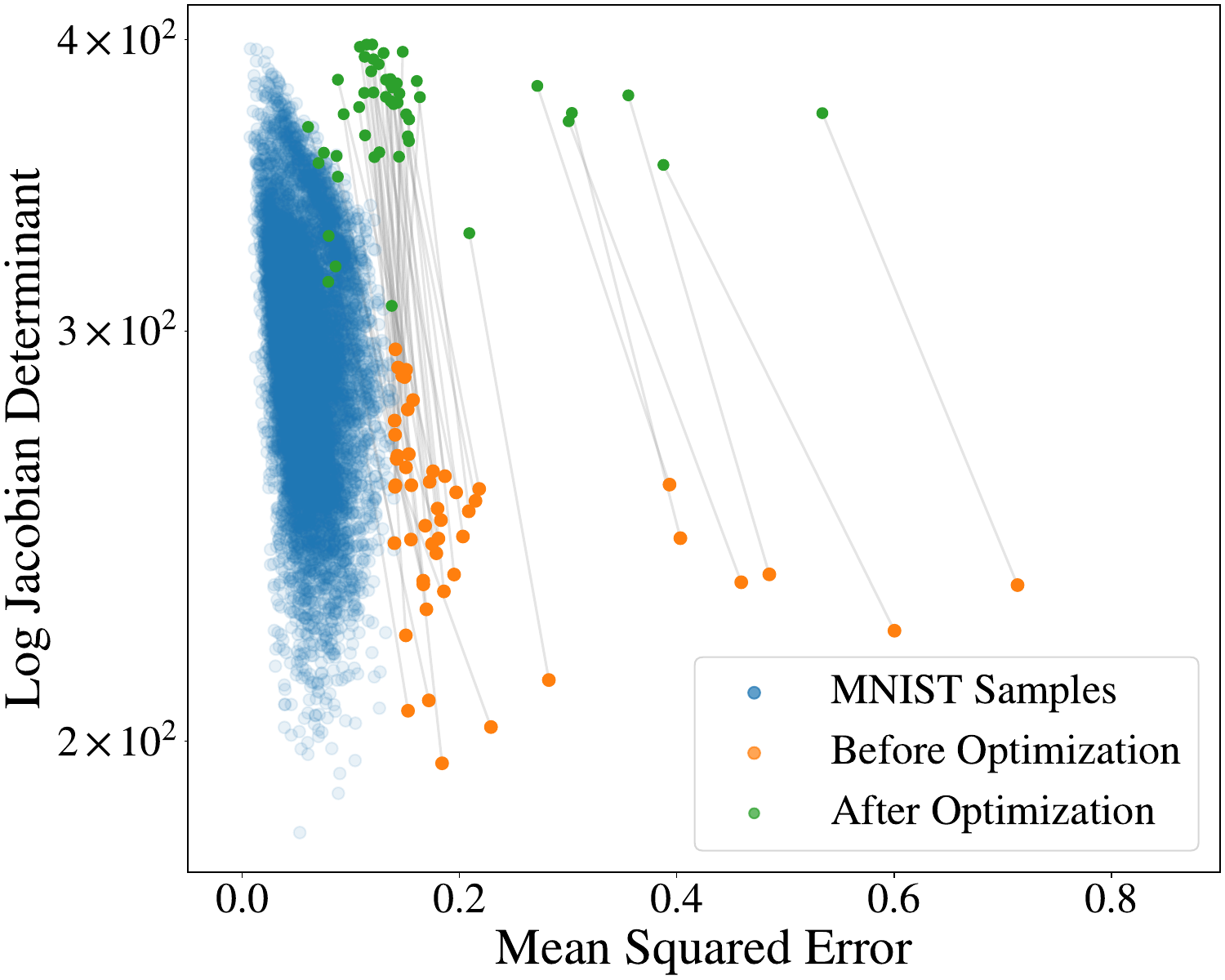}
  \caption{Improved MSE with the optimized $\mathbf{z}$ \textit{w.r.t.} log Jacobian determinant for the 50 worst MNIST samples.}
  \label{fig:likelihood2}
\end{subfigure}
\caption{Investigation of likelihood (using log Jacobian determinant as in \Cref{eq:nll}) and MSE loss with cINNs. The y-axis is plotted on a logarithmic scale for better visualization.}
\label{fig:likelihood}
\end{figure}

\subsection{Implementation Details and Preliminary Remarks}\label{sec:imp}
We compare five different approaches to optimize retinal prosthetic stimuli. The visualization of the invertible neural network architectures are depicted in \Cref{fig:a7}.

\noindent\textbf{Downsampling.} Trivial downsampling with a single learnable gain factor is applied to an input MNIST image as the visual signal. 

\noindent\textbf{Linear model.} A single linear layer is used to learn the \textit{inverse} transformation from \textit{percepts} to \textit{stimuli}. During the inference, the linear model outputs the optimized stimulus given an MNIST target image as the input. 

\noindent\textbf{Neural network.} For the \textit{non-invertible} neural network models, we use a feed-forward architecture consisting of three convolutional layers, each followed by batch normalization \cite{ioffe2015batch} and ELU activation \cite{clevert2015fast}. The network is bookended by two linear layers at the input and output stages. Similar to the linear model, the feed-forward neural network is trained to map \textit{percepts} to \textit{stimuli}.

\noindent\textbf{Invertible neural network with MMD.} 
We use the open-source library FrEIA \cite{freia} for the implementation of the invertible network.
Each INN layer is formed by a random permutation layer, Glow coupling block \cite{freia,kingma2018glow}, and an ActNorm layer \cite{kingma2018glow}. The subfunction inside the coupling block contains two ReLU-activated linear layers with a hidden dimension of 512 and outputs both $s$ and $t$ simultaneously (\Cref{fig:a7}). A stack of 9 INN layers is empirically found to be sufficient and no additional performance gains are observed beyond that. 
The affine coupling layer \cite{dinh2016density, kingma2018glow} is implemented using \texttt{GLOWCouplingBlock} in \cite{freia}. In contrast to \texttt{AllInOneBlock} in \cite{freia}, \texttt{GLOWCouplingBlock} does not include ActNorm or invertible $1\times1$ convolutions \cite{kingma2018glow}.
We also attempted to alter the dimensionality of the latent concatenations (Gaussian distribution in the forward path and zero paddings in the reverse path) either on the input or output side and applied the MMD loss in INNs, which did not result in performance improvement. 
During the training, we try to define either \textit{stimulus-to-percept} or \textit{percept-to-stimulus} as the forward process.
We compare the MSE loss of both mappings and find that defining $h$ (\Cref{eq:pdf1}) as the \textit{percept-to-stimulus} mapping makes its inverse function $h^{-1}$ easier to train, as shown in \Cref{fig:a6}, where the MSE losses are 0.2001 \textit{vs.} 0.2756 at the step of 20000. Therefore, we report the results based on this setup in \Cref{tab:1}.

\noindent\textbf{Conditional invertible neural network.} The cINN architecture follows the design of the INN, with the additional conditioning to the subfunction at each layer \cite{ardizzone2019guided,ardizzone2021conditional}. The \textit{percept} or the MNIST target is concatenated as the conditioning during the training or the inference, respectively (\Cref{fig:2right}). The conditional input $\mathbf{c}$, which is $\mathbf{y_p}$ during the training and $\tilde{\mathbf{x}}$ during the evaluation, is first transformed with a learnable linear layer to a size of $9\times9$ before being passed to the subfunction. As introduced in \Cref{sec:cinn}, the cINN is trained in an unsupervised manner only with the NLL loss (\Cref{eq:nll}).

\begin{table*}[t]
    \footnotesize
    \centering
    \caption{Reconstruction quality evaluation from the predicted percepts to the original visual signals based on the optimized stimuli. Five approaches are compared with two resolutions (Res.) of the original MNIST input image of $9\times9$ and $28\times28$ pixels. Five evaluation metrics are reported, namely Mean Absolute Error (MAE), Mean Squared Error (MSE), Structural Similarity Index Measure (SSIM), Peak Signal-to-Noise Ratio (PSNR), and classification accuracy with a pre-trained classifier (ACC). The number of the parameters of each model is listed. The top results are marked in \colorbox[HTML]{d2e3ff}{blue}, while the second-best results are marked in \colorbox[HTML]{ffd1d1}{red}.}
    \label{tab:1}
    \begin{tabular}{r c m{5em} m{5em} m{5em} m{5em} m{5em} r}
    \toprule
         \hspace{0.2em}Methods & \hspace{1em}Res. \hspace{1em} &  \hspace{0.2em}MAE $\downarrow $  & \hspace{0.2em}MSE $\downarrow $  & \hspace{0.2em}SSIM $\uparrow $ &  \hspace{0.2em}PSNR $\uparrow $ & \hspace{0.2em}ACC $\uparrow $ & \#Param $\downarrow $\\\midrule
          down  &     &   \colorbox[HTML]{ffffff}{0.1682} & \colorbox[HTML]{ffffff}{0.0619} &  \colorbox[HTML]{ffffff}{0.7248} & \colorbox[HTML]{ffffff}{30.4149} & \colorbox[HTML]{ffffff}{0.4272} & 1 \\
         linear   & 9 & \colorbox[HTML]{ffffff}{0.1657}  & \colorbox[HTML]{ffffff}{0.0584}  &  \colorbox[HTML]{ffffff}{0.7246} &  \colorbox[HTML]{ffffff}{30.6877} & \colorbox[HTML]{d2e3ff}{0.5608} & 6,561 \\
         NN  & $\times$ & \colorbox[HTML]{ffd1d1}{0.1387}  & \colorbox[HTML]{d2e3ff}{0.0436} & \colorbox[HTML]{d2e3ff}{0.7799} & \colorbox[HTML]{d2e3ff}{31.9620}  &  \colorbox[HTML]{ffd1d1}{0.5572} & 946,512 \\
         INN  & 9 &  \colorbox[HTML]{ffffff}{0.1785}  & \colorbox[HTML]{ffffff}{0.0599} & \colorbox[HTML]{ffffff}{0.7088} & \colorbox[HTML]{ffffff}{30.5656} & \colorbox[HTML]{ffffff}{0.4892} & 1,866,270 \\
         cINN  & &  \colorbox[HTML]{d2e3ff}{0.1373}  & \colorbox[HTML]{ffd1d1}{0.0570} & \colorbox[HTML]{ffd1d1}{0.7651} & \colorbox[HTML]{ffd1d1}{30.8940} & \colorbox[HTML]{ffffff}{0.5480} & 1,936,710\\\midrule
         down  &   & \colorbox[HTML]{ffffff}{0.3432}  & \colorbox[HTML]{ffffff}{0.2040} & \colorbox[HTML]{ffffff}{0.3784} & \colorbox[HTML]{ffffff}{25.0973} &  \colorbox[HTML]{ffffff}{0.2884} & 1 \\
         linear   &  28 & \colorbox[HTML]{ffffff}{0.3279}  & \colorbox[HTML]{ffffff}{0.1999} & \colorbox[HTML]{ffffff}{0.4770} & \colorbox[HTML]{ffffff}{25.1453} & \colorbox[HTML]{ffffff}{0.5884} & 63,504 \\
         NN  & $\times$ & \colorbox[HTML]{ffffff}{0.3340} & \colorbox[HTML]{ffffff}{0.1966} & \colorbox[HTML]{ffffff}{0.4901} & \colorbox[HTML]{ffffff}{25.2400} & \colorbox[HTML]{d2e3ff}{0.6192} & 4,270,900\\
         INN & 28 & \colorbox[HTML]{ffd1d1}{0.3231}  & \colorbox[HTML]{ffd1d1}{0.1947} & \colorbox[HTML]{ffd1d1}{0.4937}  & \colorbox[HTML]{ffd1d1}{25.2729} & \colorbox[HTML]{ffffff}{0.5764} & 18,063,390 \\
         cINN &  & \colorbox[HTML]{d2e3ff}{0.3136}  & \colorbox[HTML]{d2e3ff}{0.1845} & \colorbox[HTML]{d2e3ff}{0.4943} & \colorbox[HTML]{d2e3ff}{25.5161} & \colorbox[HTML]{ffd1d1}{0.5932} & 2,449,197 \\\bottomrule
    \end{tabular}
\end{table*}

\subsection{Likelihood and MSE}
We illustrate whether the maximum likelihood training in the cINN approach leads to decrease the MSE loss, as it is reported that good likelihood does not always guarantee good samples in generative models \cite{theis2015note}.
The likelihood of the optimized stimulus depends on the likelihood of the latent $\mathbf{z}$ (\Cref{eq:nll_z}). \Cref{fig:likelihood1} shows the relationship between the log-likelihood of \textit{optimized stimulus} (with $\mathbf{z}=\mathbf{0}$) and the MSE between the predicted \textit{percept} and MNIST \textit{target} (\Cref{fig:2right}). For both random and MNIST samples, the correlation between the log-likelihood and the MSE can be observed. While setting $\mathbf{z}=\mathbf{0}$ should already give a good result, we further investigate if better $\mathbf{z}\ne \mathbf{0}$ can be found to yield higher likelihood and thus lower MSE. We select the worst 50 MNIST examples with the highest MSE and modify the latent $\mathbf{z}$ to maximize the total likelihood of \textit{optimized stimulus}  using gradient descent. \Cref{fig:likelihood2} shows that the MSE values are indeed decreased for most of the samples. Further closed-loop optimization \textit{w.r.t.} $\mathbf{z}$ can be addressed in future work (\eg, \cite{guo2018closed,lohler20231}).

\subsection{Reconstruction Quality}
The visualization of the reconstruction quality with a percept resolution of $28\times28$ pixels are illustrated in \Cref{fig:vis}. We argue that negative stimulations (in red) can counteract the surrounding or elongated phosphenes, which is not applicable in downsampling. The predicted percepts (II) from the cINN show clearer turnings (\eg, the bottom left part in the image of the digit '2') and cleaner corners of the visual field. The advantages can be confirmed by quantitative assessments.
Five evaluation metrics are reported in \Cref{tab:1}, namely Mean Absolute Error (MAE), Mean Squared Error (MSE), Structural Similarity Index Measure (SSIM) \cite{wang2004image}, Peak Signal-to-Noise Ratio (PSNR), and classification accuracy with a pre-trained classifier (ACC) \cite{granley2022hybrid,wu2023embc}. These metrics encompass pixel-level measurements (MAE, MSE), fidelity-oriented assessments (SSIM, PSNR), and semantic evaluations (ACC).
The classifier is pre-trained on the MNIST dataset \cite{lecun2010mnist}, achieving a validation accuracy of 0.9921 for downsampled $9\times9$ pixel images and 0.9954 for the original $28\times28$ pixel images.
As shown in \Cref{tab:1}, cINNs demonstrate strong performance in both input resolutions, balancing pixel-level reconstruction with image-level semantic preservation.

\subsection{Resolution Discrepancies}
We experiment with two different visual signal resolutions ($9\times9$ and $28\times28$ pixels) using a fixed number of electrodes ($9\times9$). Resolution discrepancies between the stimulus and the percept indicate that retinal prosthetic optimization is fundamentally a non-bijective inverse problem.
With an increased percept resolution, cINN outperforms other approaches in various metrics. 
The number of the parameters of each model is listed in \Cref{tab:1}. The supervised models (linear model, feed-forward neural network and the MSE part in invertible neural networks) are susceptible to a higher resolution due to the fully-connected layers in the architectures, while this dense layer is only used to concatenate the \textit{target} or \textit{percept} to the subfunctions in the cINN approach. Consequently, the parameter count increases by only 1.26 times when using cINNs, compared to a 9.68-fold increase when using INNs with a corresponding 9.68-fold increase in the input size (\Cref{tab:1}).
Thanks to their inherent invertibility and generativity, both INN-MMD and cINN show robust results when the input resolution is approximately 10 times larger than the electrode array of the retinal implant. Although deep neural networks have a large number of parameters, their implementation should not be problematic on modern portable hardware, potentially leading to substantial performance improvements of reconstruction quality.

\section{Discussion}
\subsection{Limitations}
Our work relies on a pre-defined Axon Map Model with fixed $\rho$ and $\lambda$ for simulation. However, these parameters are largely patient- and implant-dependent~\cite{beyeler2022towards} and not facile to determine in real-world scenarios (\textit{cf.} \cite{fauvel2022human,granley2023human}).
Additionally, the pre-computed random samples, as illustrated in \Cref{fig:likelihood1}, are not sufficiently general, making the study of more complex datasets a challenge. Finally, we did not compare our quantitative results with those of other papers in their original forms, as they are based on different assumptions and setups, such as the computational model used in retinal prosthetic simulation and the number of electrodes in the retinal implant. Nevertheless, we contend that our experiments include a diverse range of approaches, demonstrating the proof-of-concept for leveraging INNs in retinal prosthetic optimization.

\subsection{Outlook}
We anticipate our future studies to advance in three directions. First, supervised training of INN \cite{ardizzone2018analyzing} can be further explored using different invertible architecture designs and training strategies. Second, simulation experiments can be conducted on more complex datasets utilizing other computational models to optimize stimuli. Third, eye movements of human beings, such as fixations, saccades, and micro-saccades, can be incorporated into the simulation and optimization processes.

\section{Conclusion}
We propose a stimulation optimization approach using conditional invertible neural networks for epiretinal implants and compare the reconstruction quality among downsampling, linear model, non-invertible and invertible neural networks. The cINN-based results outperform others \textit{w.r.t.} various measures, especially with a higher resolution of visual inputs and percepts. Our findings highlight the potential application with this generative, invertible, and conditional neural network to enhance the quality of retinal prostheses.

\section*{Acknowledgements}
This work was funded by the Deutsche Forschungsgemeinschaft (DFG, German Research Foundation) – grant 424556709/GRK2610.

{\small
\bibliographystyle{ieee_fullname}
\bibliography{bib.bib}

\begin{thebibliography}{10}\itemsep=-1pt

\bibitem{freia}
Lynton Ardizzone, Till Bungert, Felix Draxler, Ullrich Köthe, Jakob Kruse, Robert Schmier, and Peter Sorrenson.
\newblock {Framework for Easily Invertible Architectures (FrEIA)}, 2018-2022.

\bibitem{ardizzone2021conditional}
Lynton Ardizzone, Jakob Kruse, Carsten L{\"u}th, Niels Bracher, Carsten Rother, and Ullrich K{\"o}the.
\newblock Conditional invertible neural networks for diverse image-to-image translation.
\newblock In {\em German Conference on Pattern Recognition}, pages 373--387, 2021.

\bibitem{ardizzone2018analyzing}
Lynton Ardizzone, Jakob Kruse, Carsten Rother, and Ullrich K{\"o}the.
\newblock Analyzing inverse problems with invertible neural networks.
\newblock In {\em International Conference on Learning Representations}, 2018.

\bibitem{ardizzone2019guided}
Lynton Ardizzone, Carsten L{\"u}th, Jakob Kruse, Carsten Rother, and Ullrich K{\"o}the.
\newblock Guided image generation with conditional invertible neural networks.
\newblock {\em arXiv preprint arXiv:1907.02392}, 2019.

\bibitem{beech2024deep}
Peter Beech, Shanshan Jia, Zhaofei Yu, and Jian~K Liu.
\newblock Deep learning for visual neuroprosthesis.
\newblock {\em arXiv preprint arXiv:2401.03639}, 2024.

\bibitem{behrmann2021understanding}
Jens Behrmann, Paul Vicol, Kuan-Chieh Wang, Roger Grosse, and J{\"o}rn-Henrik Jacobsen.
\newblock Understanding and mitigating exploding inverses in invertible neural networks.
\newblock In {\em International Conference on Artificial Intelligence and Statistics}, pages 1792--1800. PMLR, 2021.

\bibitem{michael_beyeler-proc-scipy-2017}
Michael Beyeler, Geoffrey~M Boynton, Ione Fine, and Ariel Rokem.
\newblock pulse2percept: A python-based simulation framework for bionic vision.
\newblock In {\em Proceedings of the 16th Python in Science Conference}, pages 81--88, 2017.

\bibitem{beyeler2019model}
Michael Beyeler, Devyani Nanduri, James~D Weiland, Ariel Rokem, Geoffrey~M Boynton, and Ione Fine.
\newblock A model of ganglion axon pathways accounts for percepts elicited by retinal implants.
\newblock {\em Scientific Reports}, 9(1):1--16, 2019.

\bibitem{beyeler2022towards}
Michael Beyeler and Melani Sanchez-Garcia.
\newblock Towards a smart bionic eye: {AI}-powered artificial vision for the treatment of incurable blindness.
\newblock {\em Journal of Neural Engineering}, 19(6):063001, 2022.

\bibitem{Cruz19}
Edward Bloch and Lyndon da Cruz.
\newblock The {Argus} {II} retinal prosthesis system.
\newblock In {\em Prosthesis}, chapter~6. IntechOpen, 2019.

\bibitem{clevert2015fast}
Djork-Arn{\'e} Clevert, Thomas Unterthiner, and Sepp Hochreiter.
\newblock Fast and accurate deep network learning by exponential linear units {(ELUs)}.
\newblock In {\em International Conference on Learning Representations}, 2016.

\bibitem{wiki:retinal}
Wikimedia Commons.
\newblock File:retinal implant eyeimplant small.png --- wikimedia commons{,} the free media repository, 2012.
\newblock [Online; accessed 30-June-2024].

\bibitem{dinh2014nice}
Laurent Dinh, David Krueger, and Yoshua Bengio.
\newblock Nice: Non-linear independent components estimation.
\newblock {\em arXiv preprint arXiv:1410.8516}, 2014.

\bibitem{dinh2016density}
Laurent Dinh, Jascha Sohl-Dickstein, and Samy Bengio.
\newblock Density estimation using {Real NVP}.
\newblock In {\em International Conference on Learning Representations}, 2016.

\bibitem{dreher2023unsupervised}
Kris~K. Dreher, Leonardo Ayala, Melanie Schellenberg, Marco H{\"u}bner, Jan-Hinrich N{\"o}lke, Tim~J. Adler, Silvia Seidlitz, Jan Sellner, Alexander Studier-Fischer, Janek Gr{\"o}hl, Felix Nickel, Ullrich K{\"o}the, Alexander Seitel, and Lena Maier-Hein.
\newblock Unsupervised domain transfer with conditional invertible neural networks.
\newblock In {\em Medical Image Computing and Computer Assisted Intervention}, pages 770--780, 2023.

\bibitem{fauvel2022human}
Tristan Fauvel and Matthew Chalk.
\newblock Human-in-the-loop optimization of visual prosthetic stimulation.
\newblock {\em Journal of Neural Engineering}, 19(3):036038, 2022.

\bibitem{granley2021computational}
Jacob Granley and Michael Beyeler.
\newblock A computational model of phosphene appearance for epiretinal prostheses.
\newblock In {\em International Conference of the IEEE Engineering in Medicine \& Biology Society (EMBC)}, pages 4477--4481. IEEE, 2021.

\bibitem{granley2023human}
Jacob Granley, Tristan Fauvel, Matthew Chalk, and Michael Beyeler.
\newblock Human-in-the-loop optimization for deep stimulus encoding in visual prostheses.
\newblock In {\em Advances in Neural Information Processing Systems}, volume~36, pages 79376--79398, 2023.

\bibitem{granley2022hybrid}
Jacob Granley, Lucas Relic, and Michael Beyeler.
\newblock Hybrid neural autoencoders for stimulus encoding in visual and other sensory neuroprostheses.
\newblock In {\em Advances in Neural Information Processing Systems}, volume~35, pages 22671--22685, 2022.

\bibitem{greenberg1999computational}
Robert~J Greenberg, Toby~J Velte, Mark~S Humayun, George~N Scarlatis, and E De~Juan.
\newblock A computational model of electrical stimulation of the retinal ganglion cell.
\newblock {\em IEEE Transactions on Biomedical Engineering}, 46(5):505--514, 1999.

\bibitem{gretton2006kernel}
Arthur Gretton, Karsten Borgwardt, Malte Rasch, Bernhard Sch{\"o}lkopf, and Alex Smola.
\newblock A kernel method for the two-sample-problem.
\newblock {\em Advances in Neural Information Processing Systems}, 19, 2006.

\bibitem{gretton2012kernel}
Arthur Gretton, Karsten~M Borgwardt, Malte~J Rasch, Bernhard Sch{\"o}lkopf, and Alexander Smola.
\newblock A kernel two-sample test.
\newblock {\em The Journal of Machine Learning Research}, 13(1):723--773, 2012.

\bibitem{guo2013cell}
Tianruo Guo, David Tsai, John~W Morley, Gregg~J Suaning, Nigel~H Lovell, and Socrates Dokos.
\newblock Cell-specific modeling of retinal ganglion cell electrical activity.
\newblock In {\em 2013 35th Annual International Conference of the IEEE Engineering in Medicine and Biology Society (EMBC)}, pages 6539--6542. IEEE, 2013.

\bibitem{guo2018closed}
Tianruo Guo, Chih~Yu Yang, David Tsai, Madhuvanthi Muralidharan, Gregg~J Suaning, John~W Morley, Socrates Dokos, and Nigel~H Lovell.
\newblock Closed-loop efficient searching of optimal electrical stimulation parameters for preferential excitation of retinal ganglion cells.
\newblock {\em Frontiers in Neuroscience}, 12:168, 2018.

\bibitem{hartong2006retinitis}
Dyonne~T Hartong, Eliot~L Berson, and Thaddeus~P Dryja.
\newblock Retinitis pigmentosa.
\newblock {\em The Lancet}, 368(9549):1795--1809, 2006.

\bibitem{ioffe2015batch}
Sergey Ioffe and Christian Szegedy.
\newblock Batch normalization: Accelerating deep network training by reducing internal covariate shift.
\newblock In {\em International Conference on Machine Learning}, pages 448--456. PMLR, 2015.

\bibitem{isola2017image}
Phillip Isola, Jun-Yan Zhu, Tinghui Zhou, and Alexei~A Efros.
\newblock Image-to-image translation with conditional adversarial networks.
\newblock In {\em Proceedings of the IEEE Conference on Computer Vision and Pattern Recognition}, pages 1125--1134, 2017.

\bibitem{jepson2013focal}
Lauren~H Jepson, Pawe{\l} Hottowy, Keith Mathieson, Deborah~E Gunning, W{\l}adys{\l}aw D{{a}}browski, Alan~M Litke, and EJ Chichilnisky.
\newblock Focal electrical stimulation of major ganglion cell types in the primate retina for the design of visual prostheses.
\newblock {\em Journal of Neuroscience}, 33(17):7194--7205, 2013.

\bibitem{kingma2018glow}
Durk~P Kingma and Prafulla Dhariwal.
\newblock Glow: Generative flow with invertible 1x1 convolutions.
\newblock {\em Advances in Neural Information Processing Systems}, 31, 2018.

\bibitem{kingma2013auto}
Diederik~P Kingma and Max Welling.
\newblock Auto-encoding variational bayes.
\newblock In {\em International Conference on Learning Representations}, 2014.

\bibitem{kuccukouglu2022optimization}
Burcu K{\"u}{\c{c}}{\"u}ko{\u{g}}lu, Bodo Rueckauer, Nasir Ahmad, Jaap de~Ruyter van Steveninck, Umut G{\"u}{\c{c}}l{\"u}, and Marcel van Gerven.
\newblock Optimization of neuroprosthetic vision via end-to-end deep reinforcement learning.
\newblock {\em International Journal of Neural Systems}, 32(11):2250052, 2022.

\bibitem{lecun2010mnist}
Yann LeCun, L{\'e}on Bottou, Yoshua Bengio, and Patrick Haffner.
\newblock Gradient-based learning applied to document recognition.
\newblock {\em Proceedings of the IEEE}, 86(11):2278--2324, 1998.

\bibitem{lim2012age}
Laurence~S Lim, Paul Mitchell, Johanna~M Seddon, Frank~G Holz, and Tien~Y Wong.
\newblock Age-related macular degeneration.
\newblock {\em The Lancet}, 379(9827):1728--1738, 2012.

\bibitem{lohmann2019very}
Tibor~Karl Lohmann, Florent Haiss, Kim Schaffrath, Anne-Christine Schnitzler, Florian Waschkowski, Claudia Barz, Anna-Marina van Der~Meer, Claudia Werner, Sandra Johnen, Thomas Laube, et~al.
\newblock The very large electrode array for retinal stimulation {(VLARS)}—a concept study.
\newblock {\em Journal of Neural Engineering}, 16(6):066031, 2019.

\bibitem{luo2016argus}
Yvonne Hsu-Lin Luo and Lyndon Da~Cruz.
\newblock The {Argus}{\textregistered} {II} retinal prosthesis system.
\newblock {\em Progress in Retinal and Eye Research}, 50:89--107, 2016.

\bibitem{lohler2023cell}
Philipp Löhler, Andreas Albert, Andreas Erbslöh, Nruthyathi, Frank Müller, and Karsten Seidl.
\newblock A cell-type selective stimulation and recording system for retinal ganglion cells.
\newblock {\em IEEE Transactions on Biomedical Circuits and Systems}, 18(3):498--510, 2024.

\bibitem{lohler20231}
Philipp Löhler, Andreas Pickhinke, Andreas Erbslöh, and Karsten Seidl.
\newblock A 1,224-channel 60 $\mu$m pitch active closed-loop stimulator for selective retinal ganglion cell type activation.
\newblock In {\em 2023 21st IEEE Interregional NEWCAS Conference (NEWCAS)}, pages 1--4. IEEE, 2023.

\bibitem{mokwa2008intraocular}
W Mokwa, M Goertz, C Koch, I Krisch, H-K Trieu, and P Walter.
\newblock Intraocular epiretinal prosthesis to restore vision in blind humans.
\newblock In {\em International Conference of the IEEE Engineering in Medicine \& Biology Society (EMBC)}, pages 5790--5793. IEEE, 2008.

\bibitem{muandet2017kernel}
Krikamol Muandet, Kenji Fukumizu, Bharath Sriperumbudur, Bernhard Sch{\"o}lkopf, et~al.
\newblock Kernel mean embedding of distributions: A review and beyond.
\newblock {\em Foundations and Trends{\textregistered} in Machine Learning}, 10(1-2):1--141, 2017.

\bibitem{nguyen2019training}
The-Gia~Leo Nguyen, Lynton Ardizzone, and Ullrich K{\"o}the.
\newblock Training invertible neural networks as autoencoders.
\newblock In {\em German Conference on Pattern Recognition}, pages 442--455, 2019.

\bibitem{onken2016using}
Arno Onken, Jian~K Liu, PP~Chamanthi~R Karunasekara, Ioannis Delis, Tim Gollisch, and Stefano Panzeri.
\newblock Using matrix and tensor factorizations for the single-trial analysis of population spike trains.
\newblock {\em PLOS Computational Biology}, 12(11):e1005189, 2016.

\bibitem{world2019world}
World~Health Organization.
\newblock World report on vision.
\newblock 2019.

\bibitem{relic2022deep}
Lucas Relic, Bowen Zhang, Yi-Lin Tuan, and Michael Beyeler.
\newblock Deep learning--based perceptual stimulus encoder for bionic vision.
\newblock In {\em Proceedings of the Augmented Humans International Conference 2022}, pages 323--325, 2022.

\bibitem{rezende2015variational}
Danilo Rezende and Shakir Mohamed.
\newblock Variational inference with normalizing flows.
\newblock In {\em International Conference on Machine Learning}, pages 1530--1538. PMLR, 2015.

\bibitem{roessler2009implantation}
Gernot Roessler, Thomas Laube, Claudia Brockmann, Thomas Kirschkamp, Babac Mazinani, Michael Goertz, Christian Koch, Ingo Krisch, Bernd Sellhaus, Hoc~Khiem Trieu, et~al.
\newblock Implantation and explantation of a wireless epiretinal retina implant device: observations during the epiret3 prospective clinical trial.
\newblock {\em Investigative Ophthalmology \& Visual Science}, 50(6):3003--3008, 2009.

\bibitem{romeni2021machine}
Simone Romeni, Davide Zoccolan, and Silvestro Micera.
\newblock A machine learning framework to optimize optic nerve electrical stimulation for vision restoration.
\newblock {\em Patterns}, 2(7):100286, 2021.

\bibitem{sanes2015types}
Joshua~R Sanes and Richard~H Masland.
\newblock The types of retinal ganglion cells: current status and implications for neuronal classification.
\newblock {\em Annual Review of Neuroscience}, 38(1):221--246, 2015.

\bibitem{santos1997preservation}
Arturo Santos, Mark~S Humayun, Eugene de Juan, Robert~J Greenburg, Marta~J Marsh, Ingrid~B Klock, and Ann~H Milam.
\newblock Preservation of the inner retina in retinitis pigmentosa: a morphometric analysis.
\newblock {\em Archives of Ophthalmology}, 115(4):511--515, 1997.

\bibitem{sohn2015learning}
Kihyuk Sohn, Honglak Lee, and Xinchen Yan.
\newblock Learning structured output representation using deep conditional generative models.
\newblock In {\em Advances in Neural Information Processing Systems}, volume~28, 2015.

\bibitem{sorrenson2024lifting}
Peter Sorrenson, Felix Draxler, Armand Rousselot, Sander Hummerich, Lea Zimmermann, and Ullrich K{\"o}the.
\newblock Lifting architectural constraints of injective flows.
\newblock In {\em International Conference on Learning Representations}, 2024.

\bibitem{sweldens1998lifting}
Wim Sweldens.
\newblock The lifting scheme: A construction of second generation wavelets.
\newblock {\em SIAM Journal on Mathematical Analysis}, 29(2):511--546, 1998.

\bibitem{theis2015note}
Lucas Theis, A{\"a}ron van~den Oord, and Matthias Bethge.
\newblock A note on the evaluation of generative models.
\newblock In {\em International Conference on Learning Representations}, 2016.

\bibitem{tolstikhin2018wasserstein}
Ilya Tolstikhin, Olivier Bousquet, Sylvain Gelly, and Bernhard Schoelkopf.
\newblock Wasserstein auto-encoders.
\newblock In {\em International Conference on Learning Representations}, 2018.

\bibitem{tong2020stimulation}
Wei Tong, Hamish Meffin, David~J Garrett, and Michael~R Ibbotson.
\newblock Stimulation strategies for improving the resolution of retinal prostheses.
\newblock {\em Frontiers in neuroscience}, 14:262, 2020.

\bibitem{van2022end}
Jaap de~Ruyter van Steveninck, Umut G{\"u}{\c{c}}l{\"u}, Richard van Wezel, and Marcel van Gerven.
\newblock End-to-end optimization of prosthetic vision.
\newblock {\em Journal of Vision}, 22(2):20--20, 2022.

\bibitem{walter2005epiretinal}
P Walter and W Mokwa.
\newblock Epiretinal visual prostheses.
\newblock {\em Der Ophthalmologe}, 102:933--940, 2005.

\bibitem{wang2022electronic}
Bing-Yi Wang, Zhijie~Charles Chen, Mohajeet Bhuckory, Tiffany Huang, Andrew Shin, Valentina Zuckerman, Elton Ho, Ethan Rosenfeld, Ludwig Galambos, Theodore Kamins, et~al.
\newblock Electronic photoreceptors enable prosthetic visual acuity matching the natural resolution in rats.
\newblock {\em Nature communications}, 13(1):6627, 2022.

\bibitem{wang2004image}
Zhou Wang, Alan~C Bovik, Hamid~R Sheikh, and Eero~P Simoncelli.
\newblock Image quality assessment: from error visibility to structural similarity.
\newblock {\em IEEE Transactions on Image Processing}, 13(4):600--612, 2004.

\bibitem{waschkowski2014development}
Florian Waschkowski, Stephan Hesse, Anne~Christine Rieck, Tibor Lohmann, Claudia Brockmann, Thomas Laube, Norbert Bornfeld, Gabriele Thumann, Peter Walter, Wilfried Mokwa, et~al.
\newblock Development of very large electrode arrays for epiretinal stimulation {(VLARS)}.
\newblock {\em BioMedical Engineering Online}, 13(1):1--15, 2014.

\bibitem{weiland2005retinal}
James~D Weiland, Wentai Liu, and Mark~S Humayun.
\newblock Retinal prosthesis.
\newblock {\em Annual Review of Biomedical Engineering}, 7(1):361--401, 2005.

\bibitem{wong2014global}
Wan~Ling Wong, Xinyi Su, Xiang Li, Chui Ming~G Cheung, Ronald Klein, Ching-Yu Cheng, and Tien~Yin Wong.
\newblock Global prevalence of age-related macular degeneration and disease burden projection for 2020 and 2040: a systematic review and meta-analysis.
\newblock {\em The Lancet Global Health}, 2(2):e106--e116, 2014.

\bibitem{wu2023embc}
Yuli Wu, Ivan Karetic, Johannes Stegmaier, Peter Walter, and Dorit Merhof.
\newblock A deep learning-based in silico framework for optimization on retinal prosthetic stimulation.
\newblock In {\em International Conference of the IEEE Engineering in Medicine \& Biology Society (EMBC)}, pages 1--4. IEEE, 2023.

\bibitem{wu2023bmt}
Yuli Wu, Laura Koch, Peter Walter, and Dorit Merhof.
\newblock Convolutional neural network-based inverse encoder for optimization of retinal prosthetic stimulation.
\newblock {\em Biomedical Engineering / Biomedizinische Technik}, 68(s1):235, 2023.

\bibitem{zhou2022neural}
Qiongyi Zhou, Changde Du, Dan Li, Haibao Wang, Jian~K. Liu, and Huiguang He.
\newblock Neural encoding and decoding with a flow-based invertible generative model.
\newblock {\em IEEE Transactions on Cognitive and Developmental Systems}, 15(2):724--736, 2023.

\bibitem{zrenner2013fighting}
Eberhart Zrenner.
\newblock Fighting blindness with microelectronics.
\newblock {\em Science Translational Medicine}, 5(210):210ps16, 2013.

\end{thebibliography}
}

\end{document}